\documentclass[10pt,twocolumn,letterpaper]{article}
\usepackage[pagenumbers]{cvpr}
\usepackage{amsmath,amssymb}
\usepackage{booktabs}
\usepackage{array}
\usepackage{graphicx}
\usepackage{cuted}
\usepackage{caption}

\definecolor{cvprblue}{rgb}{0.21,0.49,0.74}
\usepackage[pagebackref,breaklinks,colorlinks,allcolors=cvprblue]{hyperref}
\def\confName{CVPR}
\def\confYear{2027}
\title{Ego2Flow: Benchmarking the Traffic-System Impact of Autonomous Driving on Maps Distilled from Real-World Roads}
\author{
Yu Dong\textsuperscript{1} \quad
Weiying Shen\textsuperscript{1} \quad
Hangyu Li\textsuperscript{1} \quad
Shu Han\textsuperscript{1} \quad
Hong Li\textsuperscript{2} \\[0.2em]
Xiangbo Gao\textsuperscript{3} \quad
Keshu Wu\textsuperscript{1} \quad
Zhengzhong Tu\textsuperscript{3} \quad
Soyoung Ahn\textsuperscript{1} \quad
Zhiyuan Liu\textsuperscript{2} \quad
Xiaopeng Li\textsuperscript{1,*}\\[0.5em]
{\small \textsuperscript{1}University of Wisconsin--Madison \quad
\textsuperscript{2}Southeast University \quad
\textsuperscript{3}Texas A\&M University--College Station}\quad
\small \textsuperscript{*}Corresponding author
}

\begin{document}
\maketitle
\begin{strip}
  \centering
  \includegraphics[width=\textwidth]{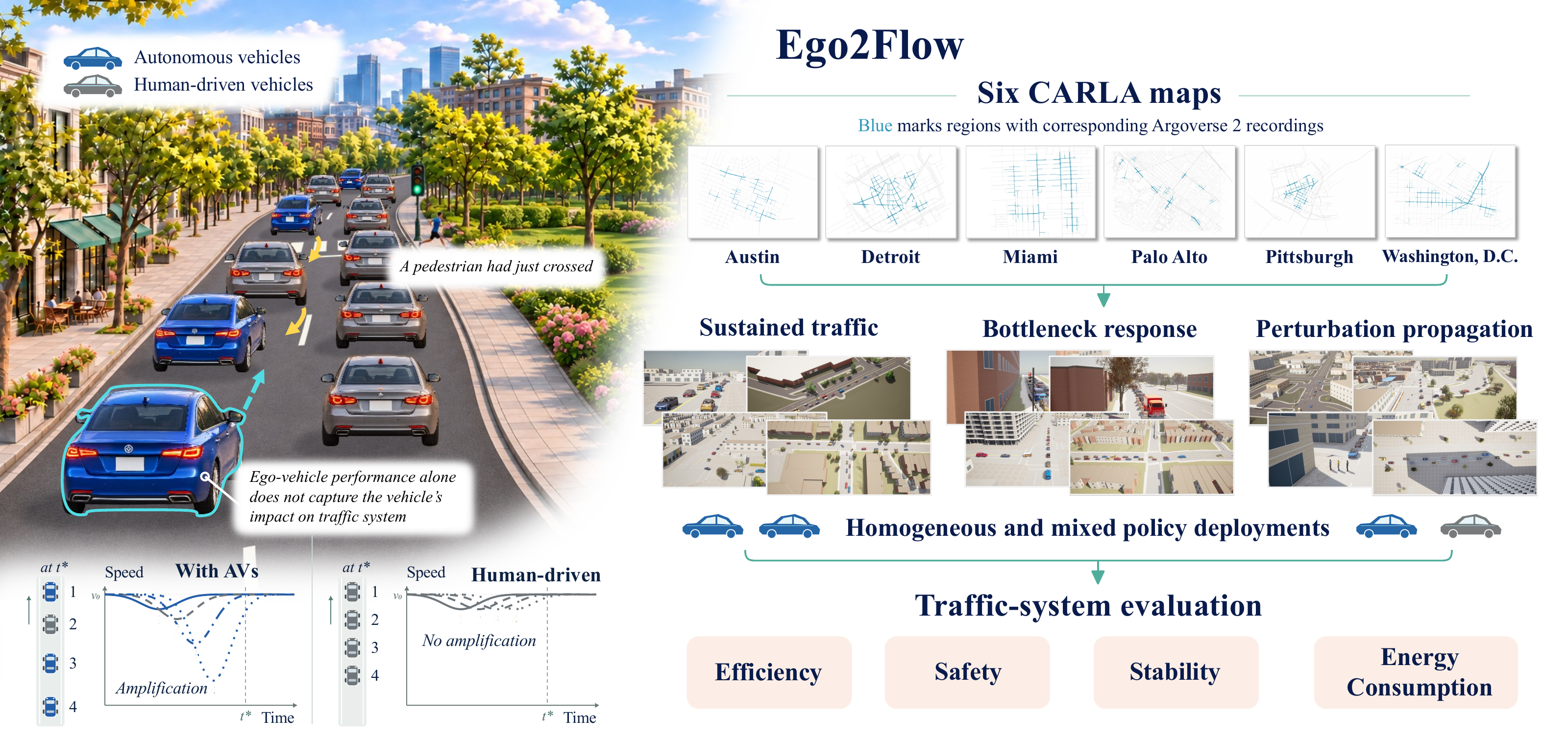}
  \captionsetup{hypcap=false}
  \captionof{figure}{\textbf{Ego2Flow: From ego-vehicle evaluation to system-level traffic-flow evaluation.}
  \emph{Motivation.} Strong ego-vehicle performance does not necessarily translate into safer or more stable traffic flow.
  The schematic illustrates one possible situation: speed disturbances may amplify along a platoon containing autonomous vehicles (AVs), increasing delays for following vehicles, while remaining unamplified in a human-driven vehicle (HV) platoon.
  Such collective effects are not captured by ego-vehicle performance alone.
  \emph{Benchmark.} Ego2Flow addresses this evaluation gap by assessing the traffic-system impact of end-to-end driving policies in closed-loop simulation.
  It provides six CARLA maps distilled from real-world roads and spatially aligned with the corresponding Argoverse~2 (AV2) regions, supporting evaluation of traffic efficiency, safety, stability and estimated energy consumption across diverse urban road environments.}
  \captionsetup{hypcap=true}
  \label{fig:overview}
\end{strip}

\begin{abstract}
End-to-end autonomous driving is predominantly evaluated through ego-vehicle safety and task completion.
Its deployment in urban traffic, however, creates sustained interactions among vehicles whose collective effects remain insufficiently evaluated.
We introduce Ego2Flow, a closed-loop benchmark for assessing the traffic-system impact of autonomous driving on maps distilled from real-world cities.
To construct realistic 3D urban environments for sustained closed-loop traffic simulation, 
we develop an agent-guided map distillation framework that integrates real-world maps and observations with complementary geographic information.
Through an iterative feedback mechanism and a traffic information graph that jointly guide scene generation,
the framework preserves lane topology, critical road geometry and decision-relevant visual conditions while simplifying nonessential details and completing unobserved regions.
Using our proposed framework, we generate and publicly release six CARLA maps from the Argoverse 2 dataset, 
covering the corresponding real-world regions and enabling paired comparisons of driving behavior under real and simulated observations.
On these maps, Ego2Flow evaluates how autonomous driving policies affect traffic efficiency, safety, stability and energy consumption across diverse road structures.
Compared with existing construction methods, our framework achieves the lowest average displacement error between planning outputs from paired real and simulated observations.
We further benchmark representative driving policies to characterize their system-level impact on urban traffic.
Project Page: https://github.com/Fisher-007/Ego2Flow.
\end{abstract}

\section{Introduction}
\label{sec:introduction}

End-to-end autonomous driving learns policies that translate sensor observations into driving decisions and has emerged as a prominent paradigm in autonomous driving research~\cite{transfuser2021,uniad2023,lead2026,onevl2026}.
Closed-loop simulation is central to evaluating these policies because it captures the feedback between their actions and subsequent observations, revealing driving behavior that open-loop prediction metrics may fail to reflect~\cite{codevilla2018offline,dauner2023misconceptions}.
Within this closed-loop setting, existing driving benchmarks primarily assess the ego vehicle's ability to navigate safely and complete its assigned task~\cite{nuplan2021,bench2drive2024,navsafe2026}.
However, urban deployment raises a further question about the traffic produced by those decisions.
A policy's ability to complete an individual trip does not establish how its widespread deployment would affect urban traffic systems.

Vehicle interactions carry the effects of a vehicle's driving decision beyond the controlled vehicle.
For example, a vehicle's braking can induce speed disturbances that propagate through a platoon and delay following vehicles.
Figure~\ref{fig:overview} illustrates this distinction: the same pedestrian crossing can induce different patterns of speed loss and spacing along a platoon, even when the leading vehicle completes the maneuver safely.
Autonomous driving policies should therefore avoid amplifying such disturbances and exacerbating delays for following vehicles.
Yet evaluating ego-vehicle performance alone cannot fully capture these collective effects.

Transferring driving control to algorithms also creates an opportunity to improve traffic-system outcomes that individual human drivers do not typically optimize for.
Learning-based control has improved traffic flow in simulation~\cite{flow2022}, and field experiments have shown that controlling a single vehicle can suppress stop-and-go waves in predominantly human-driven traffic~\cite{stern2018}.
However, experiments with commercial adaptive cruise control have also observed disturbance amplification~\cite{gunter2021string}.
These contrasting findings motivate the central question of Ego2Flow: \emph{how does a driving policy affect the traffic it joins, and does its control improve collective outcomes?}

Evaluating these effects for sensor-based driving policies requires environments that preserve decision-relevant road and visual conditions while supporting sustained interactions among vehicles.
We introduce \textbf{Ego2Flow}, a closed-loop benchmark for assessing the traffic-system impact of autonomous driving on maps distilled from real-world roads.
As summarized in Fig.~\ref{fig:overview}, it provides six CARLA~\cite{carla2017} maps covering the corresponding Argoverse~2 (AV2) regions~\cite{av22021}.
Across these diverse urban road environments, Ego2Flow assesses end-to-end driving policies through their effects on traffic efficiency, safety, stability and estimated energy consumption.

To support this evaluation, we develop an agent-guided map-distillation framework for constructing 3D urban environments.
Distillation focuses on faithfully retaining the road structure and visual cues that shape driving decisions, 
while simplifying less relevant geometry and appearance instead of reproducing every scene detail.
A traffic information graph connects these priorities to source evidence and dependencies among road elements, 
guiding the agent's decisions about how to represent observed features and complete the surrounding network (Fig.~\ref{fig:framework}). 
Static and dynamic review provide localized feedback on agreement with the recorded scene and functionality in simulation, allowing the maps to be refined for traffic evaluation.

We publicly release six CARLA maps covering the corresponding AV2 regions and connected surrounding roads, supporting joint use with AV2 data for closed-loop simulation of end-to-end driving policies. 
Across six cities, our framework achieves the lowest mean ADE between a frozen driving model's planning outputs under real and rendered observations, 9.3\% below the next-best construction method. 
Initial benchmarking of representative rule-based and learned driving policies reveals differences in traffic-system performance that task completion alone cannot capture. 
These findings establish empirical baselines and identify opportunities to improve collective traffic outcomes through driving-policy development.

Our contributions are threefold:
\begin{itemize}
    \item We introduce Ego2Flow, a closed-loop benchmark that extends the evaluation of end-to-end driving policies beyond ego-vehicle outcomes to the collective performance of urban traffic.
    \item We develop an agent-guided map-distillation framework that prioritizes information relevant to driving decisions when constructing executable 3D urban environments. A traffic information graph informs construction, while static and dynamic feedback guide refinement.
    \item We publicly release six CARLA maps spatially aligned with AV2, enabling paired evaluation of driving policies under real and simulated observations and supporting closed-loop simulation in the realistic urban regions.
\end{itemize}

\section{Related Work}
\label{sec:related}

\paragraph{End-to-end autonomous driving.}
End-to-end autonomous driving uses a unified learning framework to map raw sensor inputs directly to vehicle motion plans or control commands.
Such policies can be learned through imitation learning~\cite{lead2026} and further refined through reinforcement learning~\cite{diffusiondrivev22025}.
Recent vision--language--action (VLA) models augment this mapping with language reasoning for trajectory planning~\cite{alpamayo2025}.
World modeling can further support these policies through future scene prediction~\cite{onevl2026} and help assess navigation instructions by imagining their consequences~\cite{safedrivevla2026}.
Given the prominence of end-to-end driving, we evaluate the traffic-system effects of policies that act on sensor observations, extending beyond motion controllers operating on explicit traffic states.

\paragraph{Closed-loop evaluation.}
Closed-loop evaluation assesses driving policies through interaction, typically in simulators where actions shape subsequent observations.
Driving benchmarks emphasize safety, progress and task completion~\cite{nuplan2021,bench2drive2024,navsafe2026}, while simulation benchmarks assess the realism of agent behavior~\cite{wosac2023}.
Cooperative benchmarks examine whether perception sharing and decision negotiation improve the driving performance of connected vehicles~\cite{v2xverse2025,colmdriver2025,mdrive2026}.
At the traffic-system level, Flow~\cite{flow2022} and freeway benchmarks~\cite{mixedtraffic2026} study mixed-autonomy control and its collective effects, while IntersectionZoo~\cite{intersectionzoo2025} examines the generalization of cooperative eco-driving across urban intersections.
These traffic-control studies primarily use explicit traffic states as policy inputs in their evaluations.
Ego2Flow evaluates the traffic-system impact of sensor-based end-to-end policies in 3D urban environments that support sensor rendering and sustained vehicle interactions.

\paragraph{Environment construction for driving simulation.}
Existing pipelines construct explicit 3D environments from geographic data and scene evidence through procedural modeling and asset assembly~\cite{carla_digitaltwin,trace2026,mdrive2026}, or multimodal agent-guided generation~\cite{urbanworld22026}.
Neural reconstruction methods learn renderable scene representations from driving logs to improve visual fidelity~\cite{unisim2023,hugsim2026}, while Cam2Sim~\cite{cam2sim2026} integrates Gaussian Splatting with executable CARLA scenes.
Generative models provide complementary capabilities for synthesizing road layouts~\cite{sledge2024,scenariodreamer2025} and visual content~\cite{cosmos2025}.
We focus on distilling real-world scenes into CARLA maps that preserve driving-relevant features.
\section{Agent-Guided Map Distillation}
\label{sec:distillation}
\begin{figure*}[t]
  \centering
  \includegraphics[width=\textwidth]{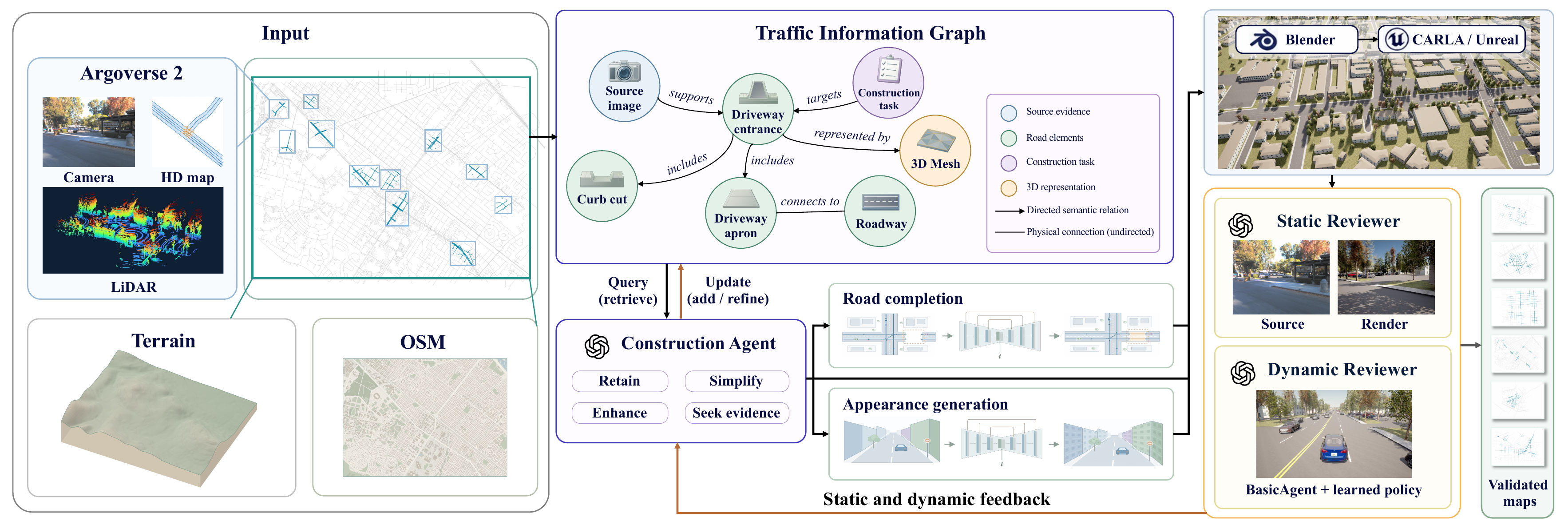}
  \caption{\textbf{Agent-guided map distillation.} 
  Aligned AV2 observations and geographic data provide the geometry and evidence for construction. 
  The traffic information graph links road elements, evidence, tasks and their simulator realizations. 
  The construction agent uses these relations to constrain road completion and appearance generation. 
  Static and dynamic reviewers return feedback for local refinement. 
  The driveway example connects a curb cut and apron to the evidence and mesh needed to preserve vehicle access.}
  \label{fig:framework}
\end{figure*}

Map distillation converts partial road observations into executable 3D environments by retaining information that governs driving and traffic interactions.
Ego2Flow assigns scene detail according to lane connectivity, physical geometry, traffic controls, and decision-relevant visibility.
As shown in Fig.~\ref{fig:framework}, a traffic information graph connects these requirements to source evidence and construction tasks.
The construction agent uses the graph to select which properties to retain and which elements to revise.
The resulting conditions guide local generation and simulator assembly, while static and dynamic feedback identify the properties requiring further revision.

\subsection{Source Alignment and Fusion}
\label{sec:source_integration}
Let $\mathcal O$ denote the available source evidence.
AV2 dataset provides local lane maps, calibrated images, LiDAR, ego poses, and object annotations~\cite{av22021}, 
OpenStreetMap supplies the surrounding road network and building footprints~\cite{osm2008} and terrain data provides regional elevation.
We register these sources in a common metric frame and fuse detailed AV2 geometry with the surrounding network.
Roads, camera poses, and actor states share the same alignment, while image timestamps and calibration retain the correspondence needed for source-aligned simulation.
The extended network supports routes and traffic interactions beyond the recorded regions.

At construction iteration $n=0,1,\ldots$, the scene representation $\mathcal S_n$ contains geometry, traffic attributes, objects, and source references.
Evidence is associated with individual properties, allowing an observed lane boundary to remain fixed while an inferred height or adjacent surface is refined.
Fusion identifies gaps between retained road segments.
The positions, tangents, widths, elevations, and lane connections at their interfaces provide boundary conditions for local completion.

\subsection{Traffic Information Graph and Construction Decisions}
\label{sec:information_graph}
\paragraph{Relational representation.}
We represent construction dependencies with an attributed graph $\mathcal G_n=(\mathcal V_n,\mathcal E_n)$, where $\mathcal V_n$ is the node set and $\mathcal E_n$ the set of typed edges.
Nodes represent road elements, evidence, construction tasks, and simulator realizations.
Edges express lane succession, evidence support, physical connections, and task dependencies.
Semantic relations are directed; physical connectivity is symmetric.
The graph associates required properties, construction decisions, and review findings with the relevant elements, while distinguishing source-supported relations from candidate associations.
The graph captures dependencies over the scene geometry in $\mathcal S_n$.

The driveway example in Fig.~\ref{fig:framework} illustrates the role of these dependencies.
Source images support the entrance, which comprises a curb cut and an apron connected to the roadway.
Preserving vehicle access requires both an opening in the curb and continuous surface elevations across the connected elements.
The graph links these requirements, their evidence, and their simulator realization to a common task.
Retrieving them together lets the agent coordinate the related modifications while preserving the entrance's traffic function.

\paragraph{From evidence to construction conditions.}
Let $A$ denote the construction agent and $F_n$ the set of available review findings, with $F_0=\varnothing$.
The agent selects tasks using the graph and feedback, then retrieves a decision context $P_n$ containing the relevant dependency subgraphs, scene geometry, source observations, prior decisions, and unresolved findings.
Retrieval follows incoming and outgoing dependencies so that adjacent structures enter the same decision context.

The agent produces a decision set $\mathcal U_n$.
Each decision identifies a target, supporting evidence, required properties, and an operation: \emph{retain}, \emph{simplify}, \emph{enhance}, or \emph{seek evidence}.
Retaining and simplifying preserve required traffic properties at different levels of detail; enhancement repairs or completes selected elements; evidence retrieval resolves uncertainty.
For example, a building can use a simplified material while retaining the geometry that occludes approaching traffic.
The decisions are recorded in the graph and translated into construction conditions $\mathcal C_n$:
\begin{equation}
\begin{aligned}
P_n &= \operatorname{Retrieve}_{A}(\mathcal G_n,F_n;\mathcal S_n,\mathcal O),\\
\mathcal U_n &= A(P_n,F_n),\\
\mathcal C_n &= \operatorname{Conditions}(\mathcal S_n,\mathcal G_n,\mathcal U_n).
\end{aligned}
\label{eq:construction_conditions}
\end{equation}
Road conditions specify the geometry to retain, editable regions, and boundary constraints.
Appearance conditions identify relevant source evidence and visual requirements.
Simulator conditions specify the physical assets and traffic controls to implement.
Unresolved properties remain associated with their tasks for further evidence gathering.
This connects traffic-relevant information selection to explicit construction constraints.

\subsection{Constrained Generation and Simulator Integration}
\label{sec:iterative_generation}
\paragraph{Local road completion.}
A pretrained Scenario Dreamer model~\cite{scenariodreamer2025} provides a geometric prior for the selected gaps.
We initialize each connection by interpolating between its source boundaries and encode it together with neighboring lanes.
Noise is applied to the editable gap latent, while retained lane latents remain fixed.
Let $z^{\rm src}$ denote the encoded lane scene before noise is added, including the initialized gap, and let $M$ be a binary mask equal to one on retained lanes and zero on the gap.
At diffusion step $t$, $\widetilde z_{t-1}$ denotes the proposed lane latent state and $z_{t-1}$ the conditioned result.
With $\odot$ denoting elementwise multiplication, each denoising step restores the retained entries:
\begin{equation}
z_{t-1}=M\odot z^{\rm src}+(1-M)\odot\widetilde z_{t-1}.
\label{eq:road_conditioning}
\end{equation}
The pretrained model weights remain fixed.
Feedback can further refine the editable latent against local geometric requirements.

During assembly, we enforce planar endpoint positions and tangents and interpolate the boundary widths and elevations.
Candidates are checked for centerline self-intersection, lane-envelope validity, and centerline intersections with buildings.
Accepted connections fill the selected gaps, while retained roads are copied directly from the source with their connections and permissions.
Latent clamping constrains sampling; source retention and boundary-constrained assembly preserve the protected geometry and its interfaces.

\paragraph{Appearance and executable map assembly.}
Cosmos~\cite{cosmos2025} generates appearance candidates from selected source image crops and the construction conditions.
The agent selects content for textures and materials according to the required properties.
Building geometry follows geographic footprints, using observed heights where available and inferred heights otherwise.
Traffic signs, markings, and controls retain verified content and placement through explicit simulator assets.
Calibrated imagery and LiDAR constrain observed access surfaces and curb profiles, preserving geometric relationships that govern occlusion and physical contact.

Let $X_n^{\rm use}$ denote accepted road geometry and $J_n^{\rm use}$ selected appearance content at iteration $n$.
Scene editing produces the updated representation $\mathcal S_{n+1}$, and assembly produces the executable CARLA map $\mathcal M_{n+1}$~\cite{carla2017}:
\begin{equation}
\begin{aligned}
\mathcal S_{n+1} &= \operatorname{Edit}(\mathcal S_n,X_n^{\rm use},\mathcal C_n),\\
\mathcal M_{n+1} &= \operatorname{Build}(\mathcal S_{n+1},\mathcal C_n,J_n^{\rm use}).
\end{aligned}
\label{eq:native_compilation}
\end{equation}
The navigation network and physical road surfaces derive from shared geometry, aligning connectivity with traversable surfaces.
Vehicles remain independently controllable, supporting both source trajectory replay and closed-loop traffic interactions.

\subsection{Static and Dynamic Feedback}
\label{sec:map_refinement}
\paragraph{Independent review.}
Static and dynamic reviewers assess $\mathcal M_{n+1}$ independently of the construction agent.
Static review compares source-aligned rendered views and geometric measurements against $\mathcal O$ and $\mathcal C_n$, examining connectivity, surface continuity, asset placement, traffic controls, visibility, and appearance.
Dynamic review follows static acceptance.
It combines route-following checks, learned-policy driving under sustained traffic, and replay of camera and actor states at source timestamps.
These tests expose navigation failures, collisions, vehicle-response problems, and interaction failures during simulation.
Each finding identifies the affected location and property together with supporting evidence.

\paragraph{Feedback-driven revision.}
Let $F_{n+1}$ collect findings from the reviews performed at this iteration.
The construction agent associates them with the corresponding elements and updates the graph to $\mathcal G_{n+1}$:
\begin{equation}
\mathcal G_{n+1}=\operatorname{Update}_{A}
\bigl(\mathcal G_n,\mathcal S_{n+1},\mathcal U_n,F_{n+1}\bigr).
\label{eq:map_feedback}
\end{equation}
The update records realized properties and retains outstanding tasks and their dependencies.
A discontinuity at a lane connection, for example, directs the agent to the linked boundary geometry and dependent surfaces.
The agent retrieves the relevant evidence, revises the construction conditions, and repeats the affected generation, assembly, and review stages.
A map is accepted once it passes both reviews, providing an environment for aligned source replay and sustained traffic-system evaluation.

\section{Ego2Flow Benchmark}
\label{sec:benchmark}

Ego2Flow measures how a driving policy changes the broader traffic system. 
The evaluation unit is the traffic generated by a set of travel requests over an observation window. 
It includes vehicles controlled by the policy, surrounding traffic and requests waiting for admission to 
the road network. Each scenario fixes the road network, demand, externally prescribed controls 
and observation window. Figure~\ref{fig:overview} organizes these evaluations into six city maps with 
three task families in two deployment settings and four metric categories. A policy can affect another 
vehicle through car-following interactions, lane occupancy or access to an intersection. Requests 
waiting for admission also experience the resulting delay. The benchmark accounts for all these 
vehicles and requests to relate driving decisions to traffic efficiency, safety, stability and 
estimated energy consumption.

\subsection{Maps and Traffic Tasks}
\label{sec:benchmark_coverage}
\label{sec:traffic_tasks}

The six maps cover Austin, Detroit, Miami, Palo Alto, Pittsburgh and Washington, D.C. 
The benchmark design allocates eight areas per city. Each city includes four areas with sustained 
traffic and two each for bottlenecks and perturbation propagation. Each area defines legal routes 
and task exits. Its traffic setup specifies initially present vehicles, incoming travel requests and 
measurement intervals. Upstream and downstream road space accommodates arrivals and queues. A buffer 
beyond each task exit allows vehicles that have completed their trips to leave without obstructing 
following traffic. 

\paragraph{Sustained traffic.}
Vehicles serve a fixed sequence of travel requests over a continuous observation window. 
Compatible AV2 observations jointly initialize vehicle positions, speeds and relative configurations. 
Additional requests specify continuing demand and scheduled release times. This task measures whether 
a policy can serve demand while maintaining safe and steady motion.

\paragraph{Bottleneck response.}
A lane restriction or a constraint on downstream discharge reduces the road's capacity to serve demand. 
The scenario fixes the arrival process as well as the restriction location and its start and end times. 
Continued arrivals reveal queue accumulation, and traffic recovery is assessed after the restriction is 
lifted. Queue and discharge time series complement the common efficiency and safety metrics.

\paragraph{Perturbation propagation.}
This task examines how vehicles respond to a disturbance transmitted through their interactions. 
After an external leader performs a sharp speed reduction, a fixed set of followers 
is observed. Neighboring traffic remains active in the simulation. Each policy determines the 
car-following behavior and lane occupancy through which the disturbance propagates. 
This task measures speed losses across all vehicles to characterize 
the traffic system's response to a perturbation under each policy.

\subsection{Deployment and Measurement}
\label{sec:benchmark_protocols}

We use two deployment settings.
Homogeneous deployment (H) assigns the tested policy to all vehicles. 
Mixed deployment (M) assigns it to a fixed subset of approximately 30\% of requests, with CARLA 
BasicAgent controlling the remaining vehicles. The external leader is controlled separately. 
H tests the collective behavior under the repeated driving policy, whereas M evaluates interactions 
with the background controller. Requests, routes, vehicle properties, 
weather, externally prescribed controls and warm-up procedures are shared across policies. Each policy 
shapes the traffic state at the start of observation through its interactions during warm-up.

We evaluate all vehicles except the external leader, clipping trajectories to the observation window 
and a region extending \(4\,\mathrm{m}\) beyond the task lane surfaces. A request is completed when 
its vehicle follows a valid route and its rear crosses the task exit. An illegal transition or 
sustained departure from the road corridor marks the request as failed, but the vehicle remains 
included in motion, safety and energy measurements. Failed requests and those awaiting admission 
accrue delay until the observation window ends, ensuring that the evaluation accounts for the full demand.

\subsection{Traffic-System Metrics}
\label{sec:system_metrics}

\paragraph{Efficiency.}
Let $\mathcal R$ denote all requests released during $[0,T)$, including those associated with 
vehicles initially present. For each request $r$, $a_r$ is its release time, set to zero for 
initially present vehicles. Let $c_r$ denote its legal completion time, with $c_r=\infty$ for 
requests not completed within the observation window. The number of legally completed requests 
is denoted by $N_{\rm completed}$.
Let \(\alpha_r(t)\) denote the ratio of forward velocity along the road tangent to the route's 
reference speed for free flow, clipped to $[0,1]$. We set $\alpha_r(t)=0$ before admission, 
during illegal motion and after route failure. The upper bound prevents speeds above the reference 
from offsetting accumulated delay. With $T$ measured in seconds, we report
\begin{equation}
 \begin{aligned}
 \bar d&=\frac{1}{|\mathcal R|}\sum_{r\in\mathcal R}
 \int_{a_r}^{\min(c_r,T)}\!\bigl(1-\alpha_r(t)\bigr)\,\mathrm dt,\\
 q_{\rm out}&=\frac{3600N_{\rm completed}}{T}.
 \end{aligned}
 \label{eq:service_metrics}
\end{equation}
The mean delay $\bar d$ measures the average time lost per request relative to travel at the reference 
speed within the observation window. It includes waiting for admission and unfinished requests. 
Throughput $q_{\rm out}$ counts completed requests per hour, including vehicles initially present.

\paragraph{Safety.}
We report collision and hard braking rates as $N_{\rm col}/D$ and $N_{\rm hb}/D$, 
where $D$ is the total evaluated distance traveled along road surfaces, in vehicle-km. 
For $N_{\rm col}$, contacts between the same two objects are counted as one collision event when gaps between consecutive contacts do not exceed \(1\,\mathrm{s}\). 
The count $N_{\rm hb}$ includes hard braking events with horizontal speed above 
$0.5\,\mathrm{m/s}$ and longitudinal acceleration below $-3\,\mathrm{m/s^2}$ for at least $0.3\,\mathrm{s}$. 

Another metric, low time-to-collision (TTC) exposure, is the time with $\mathrm{TTC}<1.5$\,s divided by total following time, 
expressed as a percentage. For positive bumper-to-bumper gap $h$ and 
closing speed $\Delta v>0.1$\,m/s, TTC equals $h/\Delta v$ and is infinite otherwise.

\paragraph{Stability.}
For stability, we measure speed variation within groups defined by road direction, occupied lane 
and maneuver. Within each group, we calculate the mean and variance of speed along the road surface, 
weighting observations by the durations they represent. Let $\sigma_g^2$ denote this variance 
and $W_g$ the observation time summed across vehicles in group $g$. 
We combine the group variances as
\begin{equation}
 \sigma_v=\left(\frac{\sum_g W_g\sigma_g^2}{\sum_g W_g}\right)^{1/2}.
 \label{eq:speed_sd}
\end{equation}
This metric captures speed variation across vehicles and over time within comparable traffic movements.

For the perturbation-propagation task, peak speed-loss ratio is calculated. Let $\mathcal B$ and $\mathcal P$ be the pre-event and 
response windows, and let $v_i(t)$ be horizontal speed. The observed peak speed-loss ratio is
\begin{equation}
 \begin{aligned}
 \bar v_i&=\frac{1}{|\mathcal B|}\int_{\mathcal B}v_i(t)\,\mathrm dt,
 &A_i&=\left[\bar v_i-\min_{t\in\mathcal P}v_i(t)\right]_+,\\
 G_{\rm obs}&=\frac{\max_{i\in\mathcal F}A_i}{A_0},
 &&A_0>0,
 \end{aligned}
 \label{eq:observed_response}
\end{equation}
where $0$ denotes the leader and $\mathcal F$ the fixed set of followers. The ratio compares 
observed speed losses, including the effects of pre-event motion and changing following behaviors.

\begin{table*}[!t]
\centering
\caption{Map-construction fidelity at a 5\,s prediction horizon. 
ADE and FDE measure differences between frozen OneVL predictions from rendered and matched real AV2 images, in metres. 
\textbf{Bold} and \underline{underlining} indicate the best and second-best values in each metric column.}
\label{tab:map_fidelity}
\label{tab:map-fidelity}
\footnotesize
\setlength{\tabcolsep}{3pt}
\begin{tabular*}{\textwidth}{@{\extracolsep{\fill}}l*{14}{r}@{}}
\toprule
\textbf{Method} & \multicolumn{12}{c}{\textbf{City-wise displacement error}} & \multicolumn{2}{c}{\raisebox{-3ex}[0pt][0pt]{\textbf{Mean}$^{\ast}$}} \\
\cmidrule(lr){2-13}
 & \multicolumn{2}{c}{Palo Alto} & \multicolumn{2}{c}{Detroit} & \multicolumn{2}{c}{Miami} & \multicolumn{2}{c}{Austin} & \multicolumn{2}{c}{Pittsburgh} & \multicolumn{2}{c}{\shortstack{Washington,\\D.C.}} & \multicolumn{2}{c}{} \\
\cmidrule(lr){2-3}\cmidrule(lr){4-5}\cmidrule(lr){6-7}\cmidrule(lr){8-9}\cmidrule(lr){10-11}\cmidrule(lr){12-13}\cmidrule(lr){14-15}
 & ADE$\downarrow$ & FDE$\downarrow$ & ADE$\downarrow$ & FDE$\downarrow$ & ADE$\downarrow$ & FDE$\downarrow$ & ADE$\downarrow$ & FDE$\downarrow$ & ADE$\downarrow$ & FDE$\downarrow$ & ADE$\downarrow$ & FDE$\downarrow$ & ADE$\downarrow$ & FDE$\downarrow$ \\
\midrule
CARLA Digital Twin Tool~\cite{carla_digitaltwin} & 0.7432 & 2.0995 & 1.1145 & 2.9851 & 1.2408 & 2.9613 & 1.2948 & 3.5720 & 1.1814 & 3.2558 & 0.9320 & 2.3710 & 1.0844 & 2.8741 \\
UrbanWorld2.0~\cite{urbanworld22026} & \underline{0.6937} & 2.0092 & 1.0826 & 2.9120 & 1.2109 & 2.9682 & 1.2728 & 3.5452 & 1.1240 & 3.1217 & 0.7031 & 1.8364 & 1.0145 & 2.7321 \\
MDrive~\cite{mdrive2026} & 0.7230 & 2.1126 & 0.7531 & 2.0016 & \underline{1.0828} & 2.6757 & 1.0315 & 2.8889 & 1.2353 & 3.5230 & 0.6074 & 1.5759 & 0.9055 & 2.4630 \\
TRACE$^{\dagger}$~\cite{trace2026} & 0.7842 & 2.2787 & 0.8755 & 2.3178 & 1.0969 & \underline{2.6450} & 1.6192 & 4.5568 & 1.1992 & 3.4441 & 0.7147 & 1.8290 & 1.0483 & 2.8452 \\
Cam2Sim$^{\ddagger}$~\cite{cam2sim2026} & 0.8632 & \underline{1.9075} & \underline{0.7275} & \textbf{1.6314} & 1.8979 & 4.0726 & \textbf{0.6161} & \textbf{1.5329} & \textbf{0.5801} & \textbf{1.5026} & \underline{0.5855} & \textbf{1.2913} & \underline{0.8784} & \textbf{1.9897} \\
\midrule
\textbf{Ego2Flow} (Ours) & \textbf{0.6095} & \textbf{1.8127} & \textbf{0.7182} & \underline{1.9024} & \textbf{1.0632} & \textbf{2.5792} & \underline{0.8117} & \underline{2.2153} & \underline{1.0482} & \underline{2.9649} & \textbf{0.5288} & \underline{1.3399} & \textbf{0.7966} & \underline{2.1357} \\
\bottomrule
\end{tabular*}
\par\smallskip
\begin{minipage}{\textwidth}
\scriptsize\raggedright
$^{\ast}$Equal-weight mean over six cities. Each city averages five areas, except
$^{\dagger}$TRACE in Pittsburgh and Washington, D.C., and $^{\ddagger}$Cam2Sim in Washington, D.C., which use four available areas due to bridge-related exclusions (TRACE) and SfM reconstruction failure (Cam2Sim). Missing cases are excluded.
\end{minipage}
\end{table*}

\paragraph{Energy.}
We estimate energy consumption at the battery terminals from simulated trajectories using a common 
longitudinal energy-balance model informed by SUMO~\cite{sumo_energy}, with fixed vehicle 
parameters from FASTSim ~\cite{fastsim2015}. For vehicle $i$, $E_{i,\rm drive}$ and $E_{i,\rm aux}$ 
denote the energy drawn from the battery for propulsion and auxiliary loads. $E_{i,\rm regen}$ denotes 
the energy returned to the battery through regenerative braking. 
Estimated net energy consumption is
\begin{equation}
 e_{\rm net}=\frac{100}{D}\sum_i
 \left(E_{i,\rm drive}+E_{i,\rm aux}-E_{i,\rm regen}\right),
 \label{eq:energy_metric}
\end{equation}
Energy and distance are accumulated over the same vehicles and trajectory intervals used for the safety metrics.

\paragraph{Total score.}
A policy may improve one aspect of traffic performance while degrading another.
We combine the eight metrics into a total score to summarize performance across
efficiency, safety, stability and energy consumption.
We normalize all eight metrics to $[0,1]$, with larger scores indicating better performance.
Throughput is normalized as
$u_q=N_{\rm completed}/|\mathcal R|$. For the other seven metrics, where smaller raw values
are better, we use $u(x;c)=\frac{1}{1+x/c}$ to prevent extreme values from dominating the aggregate. 
The fixed scales are shared across policies, areas and deployments.
\begin{center}
\small
\begin{tabular}{@{}ll@{}}
\toprule
Metric & Scale $c$ \\
\midrule
Mean delay & $60\,\text{s/request}$ \\
Collision rate & $1\,\text{event/vehicle-km}$ \\
Low-TTC exposure & $1\%$ \\
Hard braking rate & $5\,\text{events/vehicle-km}$ \\
Speed standard deviation & $2\,\mathrm{m/s}$ \\
Peak speed-loss ratio & $1$ \\
Estimated net consumption & $30\,\mathrm{kWh}/(100\,\text{vehicle-km})$ \\
\bottomrule
\end{tabular}
\end{center}
Let $\tilde u_{a,d,k}$ denote the normalized value of metric $k$ in area $a$ under deployment
$d\in\{\mathrm H,\mathrm M\}$, set to zero for anomalous runs.
We obtain the aggregate score $S_k$ by averaging the normalized values equally across deployments and areas in $\mathcal A_k$. 
The total score $S_{\rm total}$ is the equally weighted mean of the eight $S_k$.
\begin{equation}
 \begin{aligned}
 S_k&=\frac{1}{|\mathcal A_k|}\sum_{a\in\mathcal A_k}
 \frac{\tilde u_{a,\mathrm H,k}+\tilde u_{a,\mathrm M,k}}{2},\\
 S_{\rm total}&=100\times\frac{1}{8}\sum_{k=1}^{8}S_k\in[0,100].
 \end{aligned}
 \label{eq:aggregation}
\end{equation}
\section{Experiments}
\label{sec:experiments}

We evaluate Ego2Flow across six city maps to examine the consistency of planning outputs under real and rendered observations, 
the contributions of the map-distillation framework's components and the system-level effects of representative driving policies. 
Closed-loop simulations are conducted in CARLA 0.9.16~\cite{carla2017}.

\begin{table}[t]
\centering
\caption{Ablation study of the proposed framework.}
\label{tab:ablation}
\footnotesize
\setlength{\tabcolsep}{1.8pt}
\renewcommand{\arraystretch}{1.12}
\begin{tabular*}{\columnwidth}{@{\extracolsep{\fill}}lcccrr@{}}
\toprule
\textbf{Variant} & \multicolumn{3}{c}{\textbf{Framework components}} & \multicolumn{2}{c}{\textbf{Results (5\,s)}} \\
\cmidrule(lr){2-4}\cmidrule(l){5-6}
 & Graph & \shortstack{Static\\feedback} & \shortstack{Dynamic\\feedback} & \makebox[9mm][r]{ADE$\downarrow$} & \makebox[9mm][r]{FDE$\downarrow$} \\
\midrule
w/o graph & --- & $\checkmark$ & $\checkmark$ & 0.3542 & 0.9363 \\
w/o static feedback & $\checkmark$ & --- & $\checkmark$ & 0.3706 & 0.9918 \\
w/o dynamic feedback & $\checkmark$ & $\checkmark$ & --- & 0.3634 & 0.9791 \\
\midrule
\textbf{Full Ego2Flow} & $\checkmark$ & $\checkmark$ & $\checkmark$ & 0.3465 & 0.9285 \\
\bottomrule
\end{tabular*}
\end{table}

\begin{table*}[t]
\centering
\caption{System-level performance of representative driving policies on Ego2Flow.
Arrows indicate the preferred direction, and \textbf{bold} marks the best value in each column.}
\label{tab:system_results}
\footnotesize
\setlength{\tabcolsep}{2.3pt}
\renewcommand{\arraystretch}{1.12}
\begin{tabular*}{\textwidth}{@{\extracolsep{\fill}}l*{10}{r}@{}}
\toprule
\textbf{Method} & \multicolumn{2}{c}{\textbf{Efficiency}} & \multicolumn{3}{c}{\textbf{Safety}} & \multicolumn{2}{c}{\textbf{Stability}} & \textbf{Energy} & \multicolumn{2}{c}{\textbf{Overall}} \\
\cmidrule(lr){2-3}\cmidrule(lr){4-6}\cmidrule(lr){7-8}\cmidrule(lr){9-9}\cmidrule(l){10-11}
 & \shortstack{delay\\score$\uparrow$} & \shortstack{throughput\\score$\uparrow$} & \shortstack{collision\\score$\uparrow$} & \shortstack{low-TTC\\score$\uparrow$} & \shortstack{braking\\score$\uparrow$} & \shortstack{speed SD\\score$\uparrow$} & \shortstack{peak loss\\score$\uparrow$} & \shortstack{energy\\score$\uparrow$} & \shortstack{failure\\fraction$\downarrow$} & \shortstack{total\\score$\uparrow$} \\
 & \multicolumn{8}{c}{Normalized scores: $[0,1]$} & $[0,1]$ & $[0,100]$ \\
\midrule
IDM-based~\cite{idm2000} & \textbf{0.7510} & \textbf{0.8868} & 0.8733 & \textbf{0.8431} & \textbf{0.8150} & 0.5479 & \textbf{0.5242} & 0.5031 & \textbf{0.0000} & \textbf{71.804} \\
Traffic Manager~\cite{carla_tm} & 0.7048 & 0.8082 & \textbf{0.9192} & 0.5944 & 0.6038 & 0.5231 & 0.4862 & \textbf{0.5295} & \textbf{0.0000} & 64.615 \\
LEAD~\cite{lead2026} & 0.5671 & 0.5608 & 0.8360 & 0.7649 & 0.4480 & 0.5083 & 0.1528 & 0.4994 & 0.0104 & 54.215 \\
UniAD~\cite{uniad2023} & 0.4850 & 0.3349 & 0.7687 & 0.7045 & 0.3347 & \textbf{0.6107} & 0.2355 & 0.4648 & \textbf{0.0000} & 49.235 \\
Alpamayo~\cite{alpamayo2025} & 0.4550 & 0.3440 & 0.2488 & 0.6125 & 0.3082 & 0.4637 & 0.1885 & 0.3785 & 0.0208 & 37.490 \\
\bottomrule
\end{tabular*}
\par\vspace{3pt}
\begin{minipage}{\textwidth}
\scriptsize
\textit{Notes.} Failure fraction is the proportion of flagged runs among all scheduled runs per policy.
A run is flagged if the external leader's actual speed loss is below 0.2\,m/s in a perturbation-propagation scenario, 
or the total evaluated distance is below 1\,vehicle-m in a sustained-traffic or bottleneck scenario.
\end{minipage}
\end{table*}

\subsection{Map Fidelity Evaluation}
\label{sec:map_fidelity}

Across the six city maps, we compare our framework with the CARLA Digital Twin Tool~\cite{carla_digitaltwin}, UrbanWorld2.0~\cite{urbanworld22026}, MDrive~\cite{mdrive2026}, TRACE~\cite{trace2026} and Cam2Sim with Gaussian Splatting (GS)~\cite{cam2sim2026,gaussian2023}. 
For each recorded observation window, the reconstructed scene reproduces the camera pose, calibration and surrounding traffic states in the corresponding AV2 recordings. 
A frozen OneVL model~\cite{onevl2026} receives paired real and rendered observations with the same context available up to the prediction time and the same inference settings. 
Let $\hat{\mathbf p}^{\rm real}_j$ and $\hat{\mathbf p}^{\rm sim}_j$ be its predicted positions at the same future time points, expressed in the same ego-vehicle coordinate frame at the reference time. 
For $K$ samples ending at 5\,s, the average displacement error (ADE) and final displacement error (FDE) are
\begin{equation}
 \begin{aligned}
 \mathrm{ADE}&=\frac{1}{K}\sum_{j=1}^{K}
 \left\|\hat{\mathbf p}^{\rm sim}_j-\hat{\mathbf p}^{\rm real}_j\right\|_2,\\
 \mathrm{FDE}&=\left\|\hat{\mathbf p}^{\rm sim}_K-\hat{\mathbf p}^{\rm real}_K\right\|_2.
 \end{aligned}
 \label{eq:paired_fidelity}
\end{equation}

Table~\ref{tab:map_fidelity} reports city-wise results and their overall means.
Ego2Flow ranks first or second in both ADE and FDE across all six cities and in the overall averages.
Where Ego2Flow ranks second, Cam2Sim achieves the lowest error.
Despite variation in baseline performance across cities, Ego2Flow maintains consistently strong planning-output fidelity, achieving the lowest mean ADE of 0.7966\,m, 9.3\% below Cam2Sim.
Notably, Cam2Sim incorporates Gaussian Splatting to reconstruct recorded appearance, while Ego2Flow achieves these competitive results using native CARLA rendering.
These findings support our distillation strategy of prioritizing driving-relevant structure and visual cues, demonstrating that the resulting maps preserve the evaluated model's planning outputs across diverse urban environments.

\paragraph{Framework Component Analysis}
\label{sec:ablation_experiments}

To assess the contribution of each framework component, we conduct an ablation study on a randomly selected scene from the Palo Alto map.
We compare the full framework with variants that individually omit the traffic information graph, static feedback or dynamic feedback, using the same paired OneVL evaluation.
As shown in Table~\ref{tab:ablation}, the full framework achieves the lowest ADE and FDE.
Removing static feedback produces the largest ADE increase of 7.0\%, while separately omitting dynamic feedback and the traffic information graph increases ADE by 4.9\% and 2.2\%, respectively.
These comparisons support the effectiveness of each component in the proposed framework.

\subsection{Traffic-System Policy Evaluation}
\label{sec:traffic_experiments}

We compare two rule-based baselines, CARLA Traffic Manager (TM)~\cite{carla_tm} and an IDM-based controller~\cite{idm2000}, 
with three end-to-end policies: UniAD~\cite{uniad2023}, Alpamayo~\cite{alpamayo2025} and LEAD~\cite{lead2026}.
The rule-based baselines use simulator state, while the end-to-end policies act on sensor observations with fixed model weights.
Closed-loop simulations use synchronous 0.05\,s time steps and identical vehicle physics settings across policies.

Table~\ref{tab:system_results} reports normalized system-level scores aggregated across the applicable scenarios and both deployment settings.
The IDM-based controller achieves the highest total score of 71.804, followed by TM at 64.615.
LEAD performs best among the evaluated end-to-end policies, with a total score of 54.215.
However, the overall ranking conceals differences across individual metrics, with no policy achieving the best score in every dimension.

\paragraph{Performance across metrics.}
Traffic efficiency, safety, stability and energy consumption capture different aspects of system-level performance.
A policy need not achieve the best value in every metric and policy development can instead seek Pareto-efficient trade-offs that reflect deployment priorities.
The IDM-based controller leads in efficiency, low-TTC, braking and peak loss scores, while TM achieves the best collision and energy scores.
Among the end-to-end policies, UniAD achieves the highest speed SD score among all evaluated policies. 
Compared with LEAD, it scores higher on stability metrics but lower in the other dimensions.
These differences reveal strengths and weaknesses that individual metrics cannot fully describe, providing directions for policy improvement beyond ego-vehicle performance.

\paragraph{Overall performance.}
The total score summarizes performance across metrics, while the failure fraction records the proportion of runs that fail the evaluation checks, including those with negligible vehicle movement or insufficient leader speed loss.
The rule-based baselines lead most of the individual metric rankings and achieve higher total scores than the evaluated end-to-end policies.
These results do not negate their reported ego-level driving capabilities, for example, LEAD achieves strong performance on ego-level closed-loop benchmarks such as Bench2Drive~\cite{lead2026}.
The contrast suggests that strong ego-level performance does not necessarily translate into better traffic-system outcomes.
Jointly assessing ego-level and system-level performance can therefore reveal limitations that individual-vehicle evaluation may overlook and guide policy development toward improved urban traffic performance.
\section{Conclusion}
\label{sec:conclusion}

We presented Ego2Flow, a closed-loop benchmark for assessing how end-to-end driving policies affect the traffic they join.
Our agent-guided map-distillation framework constructs connected 3D environments by preserving driving-relevant road structure and visual cues while simplifying nonessential detail.
Paired evaluations demonstrate strong planning-output fidelity, with the lowest mean ADE among the compared construction methods.
Benchmarking representative policies across 48 scenarios reveals trade-offs among traffic efficiency, safety, stability and energy consumption, 
showing that more sophisticated driving models do not necessarily yield better traffic-flow performance.
We publicly release six CARLA maps spatially aligned with AV2, supporting paired evaluation and closed-loop simulation in the corresponding urban regions.
These resources and empirical findings provide a basis for developing driving policies that account for their effects on surrounding traffic and improve collective outcomes.
\label{maintextend}
{
\small
\bibliographystyle{ieeenat_fullname}
\bibliography{main}
}
\end{document}